\documentclass[11pt]{article}

 \usepackage{natbib}
 \usepackage{amssymb}
 \usepackage{amsbsy}
 \usepackage{amscd}
 \usepackage{amsmath}
 \usepackage{amsthm}

\newtheorem{Theorem}{{\bf Theorem}}
\newtheorem{Lemma}{{\bf Lemma}}
\newtheorem{Corollary}{{\bf Corollary}}
\newtheorem*{keywords}{\bf Keywords}
\newcommand{\RR}{\mathbb R}

\newcommand{\EE}{\mathbb E}

\title{A Widely Applicable Bayesian Information Criterion}

\author{
Sumio Watanabe \\
Department of Computational Intelligence and Systems Science\\
Tokyo Institute of Technology\\
Mailbox G5-19, 4259 Nagatsuta, Midori-ku\\
Yokohama, Japan 226-8502 \\
swatanab@dis.titech.ac.jp \\
}

\begin{document}

\maketitle

\begin{abstract}
A statistical model or a learning machine is called regular if the map taking a parameter 
to a probability distribution is one-to-one and if its Fisher information 
matrix is always positive definite. If otherwise, it is called singular.
In regular statistical models, 
the Bayes free energy, which is defined by the minus logarithm of Bayes marginal likelihood, 
can be asymptotically approximated by the 
Schwarz Bayes information criterion (BIC), whereas in singular models
such approximation does not hold. 

Recently, it was proved that the Bayes free energy of a singular model is
asymptotically given by a generalized formula
using a birational invariant, the real log canonical threshold (RLCT),
instead of half the number of parameters in BIC. 
Theoretical values of RLCTs in several statistical models are now being 
discovered based on algebraic geometrical methodology. 
However, it has been difficult to estimate the Bayes free energy using only training samples, 
because an RLCT depends on an unknown true distribution. 

In the present paper, we define a widely applicable Bayesian information criterion (WBIC) by 
the average log likelihood function 
over the posterior distribution with the inverse temperature $1/\log n$,
where $n$ is the number of training samples. We mathematically prove that 
WBIC has the same asymptotic expansion as the Bayes free energy, even if
a statistical model is singular for and  unrealizable by a statistical model. 
Since WBIC can be numerically calculated without any information about a true 
distribution, 
it is a generalized version of BIC onto singular statistical models. 
\end{abstract}

\begin{keywords}
Bayes marginal likelihood, Widely applicable Bayes Information Criterion
\end{keywords}

\section{Introduction}

A statistical model or a learning machine is called regular if the map taking a parameter 
to a probability distribution is one-to-one and if its Fisher information 
matrix is always positive definite. If otherwise, it is called singular.
Many statistical models and learning machines are not regular but singular,
for example, artificial neural networks, normal mixtures, binomial mixtures, 
reduced rank regressions, Bayesian networks, and hidden Markov models. 
In general, if a statistical model
contains hierarchical layers, hidden variables, or grammatical
rules, then it is singular. In other words, if a statistical model is devised so that 
it extracts hidden structure from a random phenomenon, then it naturally becomes 
singular. 
If a statistical model is singular, then 
the likelihood function cannot be approximated by any normal distribution, 
resulting that neither AIC, BIC, nor MDL can be used in statistical 
model evaluation. Hence 
constructing singular learning theory is an important issue 
in both statistics and learning theory.

A statistical model or a learning machine is represented by  a probability density function
$p(x|w)$ of $x\in{\RR}^{N}$ for a given parameter $w\in W\subset{\RR}^{d}$,
where $W$ is a set of all parameters. 
A prior probability density function is denoted by $\varphi(w)$ on $W$. 
Assume that training samples $X_{1},X_{2},...,X_{n}$ are independently 
subject to a probability density function $q(x)$, which is called 
a true distribution. 
The log loss function or the minus log likelihood function is defined by 
\begin{equation}\label{eq:Ln(w)}
L_{n}(w)=-\frac{1}{n}\sum_{i=1}^{n}\log p(X_{i}|w).
\end{equation}
Also the Bayes free energy ${\cal F}$ is defined by 
\begin{equation}\label{eq:FFF}
{\cal F}= -\log \int \prod_{i=1}^{n}p(X_{i}|w)\varphi(w)dw.
\end{equation}
This value ${\cal F}$ can be understood as the minus logarithm of marginal likelihood of a model and a prior,
hence it plays an important role in statistical model evaluation. In fact, 
a model or a prior is often optimized by maximization of the Bayes marginal likelihood \citep{Good}, 
which is equivalent to minimization of the Bayes free energy. 

If a statistical model is regular, then the posterior distribution 
can be asymptotically approximated by a normal distribution, resulting that
\begin{equation}\label{eq:BIC0}
{\cal F} \cong nL_{n}(\hat{w})+\frac{d}{2}\log n,
\end{equation}
where $\hat{w}$ is the maximum likelihood estimator, 
$d$ is the dimension of the parameter space, and $n$ is the number of training 
samples. The right hand side of eq.(\ref{eq:BIC0}) 
is the well-known Schwarz Bayesian information criterion (BIC) \citep{Schwarz}. 

If a statistical model is singular, then the posterior distribution 
is different from any normal distribution, hence the Bayes free energy 
cannot be approximated by BIC in general. 
Recently, it was proved in \citep{LNCS1999,NC2001,Cambridge2009,IWSMI2010} that, 
even if a statistical model is singular, 
\[
{\cal F}\cong nL_{n}(w_{0})+\lambda \log n,
\]
where $w_{0}$ is the parameter that minimizes the Kullback-Leibler 
distance from a true distribution to a statistical model, and 
$\lambda>0$ is a rational number called the real log canonical threshold (RLCT).

The birational invariant RLCT, which was firstly found by a research of singular Schwartz distribution \citep{Gelfand},
plays an important role in algebraic geometry and algebraic analysis \citep{Bernstein,Sato,Kashiwara,Varchenko,Kollar,Saito}. 
In algebraic geometry, it represents a relative property of singularities of a pair of algebraic varieties. 
In statistical learning theory, it shows 
the asymptotic behaviors of the Bayes free energy and the generalization loss,
which are determined by a pair of an optimal parameter set and a parameter set $W$.  

If a set of a true distribution, a statistical model, and a prior distribution are fixed, 
then there is an algebraic geometrical procedure which enables us to find an RLCT \citep{Hironaka}.
In fact, RLCTs for several statistical models and learning machines are being 
discovered. For example, RLCTs have been studied in 
artificial neural networks \citep{NN2001,Aoyagi2}, normal mixtures \citep{Yamazaki}, 
reduced rank regressions \citep{Aoyagi}, 
Bayes networks \citep{Rusakov,Zwiernik,Zwiernik2}, 
binomial mixtures, Boltzmann machines \citep{Yamazaki2}, and hidden Markov models. To study 
singular statistical models, new 
algebraic geometrical theory is constructed \citep{Cambridge2009,Drton,Lin,Kiraly}. 

Based on such researches, theoretical behavior of the Bayes free energy 
is clarified. These results are very important because they 
indicate the quantitative difference of singular models from regular ones. 
However, in general, an RLCT
depends on an unknown true distribution.
In practical applications, we do not know a 
true distribution, hence we cannot directly apply the 
theoretical results to statistical model evaluation. 

In the present paper, in order to estimate the Bayes free energy 
without any information about a true distribution, we propose 
a widely applicable Bayesian information criterion (WBIC) by 
the following definition.
\begin{equation}\label{eq:WBICDEF}
\mathrm{WBIC}=\EE_{w}^{\beta}[nL_{n}(w)],\;\;\;\beta=\frac{1}{\log n},
\end{equation}
where $\EE_{w}^{\beta}[\;\;]$ shows the expectation value 
over the posterior distribution on $W=\{w\}$
that is defined by, for an arbitrary integrable function $G(w)$,
\begin{equation}\label{eq:BayesP}
\EE_{w}^{\beta}[G(w)] = \frac{\displaystyle
\int G(w)\;\prod_{i=1}^{n}p(X_{i}|w)^{\beta}\;\varphi(w)dw
}{\displaystyle
\int \prod_{i=1}^{n}p(X_{i}|w)^{\beta}\;\varphi(w)dw
}.
\end{equation} 
In this definition, $\beta>0$ is called the inverse temperature. 
Then the main purpose of this paper is to show
\[
{\cal F}\cong \mathrm{WBIC}.
\]
To establish mathematical support of WBIC, we prove 
three theorems. 
Firstly, in Theorem \ref{Theorem:unique} we show that
there exists a unique inverse temperature $\beta^{*}$ which satisfies
\[
{\cal F}=\EE_{w}^{\beta^{*}}[nL_{n}(w)].
\]
The optimal inverse temperature $\beta^{*}$ satisfies the convergence
in probability, $\beta^{*}\log n\rightarrow 1$ as $n\rightarrow\infty$. 
Secondly, in Theorem \ref{Theorem:main} we prove that, 
even if a statistical model is 
singular, 
\[
\mathrm{WBIC}\cong nL_{n}(w_{0})+\lambda\log n.
\]
In other words, WBIC has the same asymptotic behavior as the 
Bayes free energy even if a statistical model is singular. 
And lastly, in Theorem \ref{Theorem:regular} we prove that, 
if a statistical model is regular, then 
\[
\mathrm{WBIC}\cong nL_{n}(\hat{w})+\frac{d}{2}\log n,
\]
which shows WBIC coincides with BIC in regular statistical models. 
Moreover, a computational cost in numerical calculation of WBIC is 
far smaller than that of the Bayes free energy. 
These results show that WBIC is a generalized version of BIC onto
singular statistical models and that RLCTs can be estimated 
even if a true distribution is unknown. 

This paper consists of eight sections. 
In Section 2, we summarize several notations. In Section 3, singular
learning theory and standard representation theorem are introduced. 
The 
main theorems and corollaries of this paper are explained in Section 4, which are 
mathematically proved in Section 5. As the purpose of the present paper
is to prove the mathematical support of WBIC, Sections 4 and 5 are
the main sections. 
In section 6, a method how to use WBIC in statistical model evaluation is 
illustrated using an experimental result.
In section 7 and 8, we discuss and conclude the present paper.

\begin{table}[tb]
\begin{center}
\begin{tabular}{|c|c|c|}
\hline
Variable & Name & eq. number \\
\hline
${\cal F}$ & Bayes free energy & eq.(\ref{eq:FFF}) \\
${\cal G}$ & Generalization loss & eq.(\ref{eq:GGG}) \\
\hline
$\mathrm{WBIC}(n)$ & WBIC
&eq.(\ref{eq:WBICDEF})\\
$\mathrm{WAIC}(n)$ & WAIC
&eq.(\ref{eq:WAICDEF})\\
\hline
$\EE_{w}^{\beta}[\;\;]$ & posterior average &eq.(\ref{eq:BayesP}) \\
$\beta^{*}$ & optimal inverse temperature & eq.(\ref{eq:unique})\\
\hline 
$L(w)$ & log loss function &eq.(\ref{eq:L(w)})\\
$L_{n}(w)$ & empirical loss & eq.(\ref{eq:Ln(w)}) \\
\hline 
$K(w)$ & Kullback-Leibler distance & eq.(\ref{eq:K(w)})\\
$K_{n}(w)$ & empirical KL distance & eq.(\ref{eq:Kn}) \\
\hline
$\lambda$ & real log canonical threshold & eq.(\ref{eq:RLCT}) \\
$m$ & multiplicity & eq.(\ref{eq:mmm})\\
$Q(K(w),\varphi(w))$ & parity of model &  eq.(\ref{eq:QQQ}) \\
\hline
$({\cal M},g(u),a(x,u),b(u))$ & resolution quartet & Theorem \ref{Theorem:base} \\
\hline
\end{tabular}
\end{center}
\caption{Variable, Name, and Equation Number}
\label{table:111}
\end{table}

\section{Statistical Models and Notations} 

In this section, we summarize several notations. 
Table \ref{table:111} shows variables, names, and equation numbers
in this paper. 
The average log loss function $L(w)$ and the entropy of the true distribution $S$ 
are respectively defined by
\begin{eqnarray}
L(w)&=&-\int q(x) \log p(x|w)dx, \label{eq:L(w)}\\
S&=&-\int q(x)\log q(x)dx. 
\end{eqnarray}
Then $L(w)=S+D(q||p_{w})$, where $D(q||p_{w})$ is 
the Kullback-Leibler distance defined by 
\[
D(q||p_{w})=\int q(x)\log \frac{q(x)}{p(x|w)}dx.
\]
Then $D(q||p_{w})\geq 0$, hence $L(w)\geq S$. Moreover, $L(w)=S$ 
if and only if $p(x|w)=q(x)$. 

In this paper, 
we assume that there exists a parameter $w_{0}$ in the open kernel of
$W$ which minimizes $L(w)$, 
\[
L(w_{0})=
\min_{w\in W} L(w),
\]
where the open kernel of a set $S$ is the defined by the 
largest open set that is contained in $S$.
Note that such $w_{0}$ is not unique in general, because the map 
$w\mapsto p(x|w)$ is not one-to-one in general in singular statistical models. 
We also assume that, for an arbitrary $w$ that satisfies $L(w)=L(w_{0})$, $p(x|w)$ is 
the same probability density function. Let $p_{0}(x)$ be such a unique probability 
density function. In general, the set
\[
W_{0}=\{w\in W; p(x|w)=p_{0}(x)\}
\]
is not a set of single element but 
an analytic set or an algebraic set with singularities. 
Let us define a log density ratio function, 
\[
f(x,w)=\log\frac{p_{0}(x)}{p(x|w)},
\]
which is equivalent to 
\[
p(x|w)=p_{0}(x)\exp(-f(x,w)).
\]
Two functions $K(w)$ and $K_{n}(w)$ are respectively defined by
\begin{eqnarray}
K(w)&=&\int q(x)f(x,w)dx, \label{eq:K(w)}\\
K_{n}(w)&=&\frac{1}{n}\sum_{i=1}^{n}f(X_{i},w).\label{eq:Kn}
\end{eqnarray}
Then it immediately follows that
\begin{eqnarray}
L(w)&=&L(w_{0})+K(w),\\
L_{n}(w)&=&L_{n}(w_{0})+K_{n}(w).
\end{eqnarray}
The expectation value over all sets of training samples 
$X_{1},X_{2},...,X_{n}$ is denoted by
$\EE[\;\;]$. For example, $\EE[L_{n}(w)]=L(w)$ and $\EE[K_{n}(w)]=K(w)$. 
The problem of statistical learning is characterized by the 
log density ratio function $f(x,w)$. In fact, 
\begin{eqnarray}
\EE_{w}^{\beta}[nL_{n}(w)]&=&nL_{n}(w_{0})+\EE_{w}^{\beta}[nK_{n}(w)], \label{eq:LnKn} \\
\EE_{w}^{\beta}[nK_{n}(w)]&=&\frac{\int nK_{n}(w)\exp(-n\beta K_{n}(w))\varphi(w)dw}
{\int \exp(-n\beta K_{n}(w))\varphi(w)dw}.\label{eq:KnKn}
\end{eqnarray}
The main purpose of the present paper is to prove
\[
{\cal F}\cong nL_{n}(w_{0})+\EE_{w}^{\beta}[nK_{n}(w)].
\]
for $\beta=1/\log n$. 
\vskip5mm\noindent
{\bf Definition}. \\
(1) If 
$q(x)=p_{0}(x)$, then $q(x)$ is said to be {\it realizable} by $p(x|w)$.
If otherwise, it is said to be {\it unrealizable}.\\
(2) If the set $W_{0}$ consists of a single element $w_{0}$ and if 
the Hessian matrix
\begin{equation}\label{eq:Jij}
J_{ij}(w)=\frac{\partial^{2} L}{\partial w_{i}\partial w_{j}}(w)
\end{equation}
at $w=w_{0}$ is strictly positive definite, 
$q(x)$ is said to be {\it regular} for $p(x|w)$. 
If otherwise, then it is said to be {\it singular} for $p(x|w)$. 
\vskip5mm\noindent
Note that the matrix $J(w)$ is equal to the Hessian matrix of $K(w)$ 
and that $J(w_{0})$ is equal to the Fisher information matrix if the true distribution
is realizable by a statistical model.

\section{Singular Learning Theory}

In this section we summarize singular learning theory. In the present paper, 
we assume the following conditions. 
\vskip5mm\noindent
{\bf Fundamental Conditions.} \\
(1) The set of parameters $W$ is a compact set in ${\RR}^{d}$ whose open
kernel is not the empty set. 
Its boundary is defined by several analytic functions, in other 
words,
\[
W=\{w\in {\RR}^{d};\pi_{1}(w)\geq 0,\pi_{2}(w)\geq 0,...,
\pi_{k}(w)\geq 0\}.
\]
(2) The prior distribution satisfies $\varphi(w)=\varphi_{1}(w)\varphi_{2}(w)$, 
where $\varphi_{1}(w)\geq 0$ is an analytic function and $\varphi_{2}(w)>0$ is
a $C^{\infty}$-class function. \\
(3) Let $s\geq 6$ and 
\[
L^{s}(q)=\{f(x);\|f\|_{s}\equiv \Bigl(\int |f(x)|^{s}q(x)dx\Bigr)^{1/s}<\infty\}
\]
be a Banach space. There exists an open set $W'\supset W$ such that the map $W'\ni w\mapsto f(x,w)$ is 
an $L^{s}(q)$-valued analytic function. \\
(4) The set $W_{\epsilon}$ is defined by
\[
W_{\epsilon}=\{w\in W\;;\;K(w)\leq \epsilon\}.
\]
It is assumed that there exist constants $\epsilon,c>0$ such that
\begin{equation}\label{eq:finite}
(\forall w\in W_{\epsilon})\;\;\;
\EE_{X}[f(X,w)]\geq c\;\EE_{X}[f(X,w)^{2}].
\end{equation}

\noindent{\bf Remark.} 
(1) These conditions allow that
the set of optimal parameters 
\[
W_{0}=\{w\in W\;;\;p(x|w)=p(x|w_{0})\}
=\{w\in W\;;\;K(w)=0\}
\]
may contain singularities, and that
the Hessian matrix $J(w)$ at $w\in W_{0}$ is not 
positive definite. Therefore $K(w)$ can not be approximated 
by any quadratic form in general. \\
(2) The condition eq.(\ref{eq:finite}) is satisfied if a true distribution is realizable by
or regular for a statistical model \citep{IWSMI2010}. If a true distribution is
unrealizable by and singular for a statistical model, this condition is not satisfied
in general. In the present paper, we study the case when eq.(\ref{eq:finite}) is satisfied.

\begin{Lemma}\label{Lemma:start}
Assume Fundamental Conditions (1)-(4). Let
\[
\beta=\frac{\beta_{0}}{\log n}, 
\]
where $\beta_{0}>0$ is a constant and let $0\leq r<1/2$. Then, as $n\rightarrow \infty$, 
\begin{eqnarray}
&&\int_{K(w)\geq 1/n^{r}}\exp(-n\beta K_{n}(w))\varphi(w)dw =o_{p}(\exp(-\sqrt{n})),\label{eq:r11} \\
&&\int_{K(w)\geq 1/n^{r}}nK_{n}(w)\exp(-n\beta K_{n}(w))\varphi(w)dw=o_{p}(\exp(-\sqrt{n})).\label{eq:r12}
\end{eqnarray}
\end{Lemma}
The proof of Lemma \ref{Lemma:start} is given in Section \ref{section:proof}.
\vskip5mm\noindent
Let $\epsilon>0$ be a sufficiently small constant. 
Lemma \ref{Lemma:start} shows that integrals outside of the region $W_{\epsilon}$ do not
affect the expectation value $\EE_{w}^{\beta}[nK_{n}(w)]$ asymptotically, because 
in the following theorems, we prove that integrals in the region $W_{\epsilon}$ have
larger orders than them. 
To study integrals in the region $W_{\epsilon}$, we need algebraic geometrical method,
because the set $\{w;K(w)=0\}$ contains singularities in general. There are quite many 
kinds of singularities, however, 
the following theorem makes any singularities be a same standard form. 

\begin{Theorem}(Standard Representation) \label{Theorem:base}
Assume Fundamental Conditions (1)-(4). Let $\epsilon>0$ be a sufficiently small constant. 
Then there exists an quartet $({\cal M}, g(u), a(x,u), b(u))$, where \\
(1) ${\cal M}$ is a $d$ dimensional real analytic manifold, \\
(2) $g$ is a proper analytic function $g:{\cal M}\rightarrow W_{\epsilon}'$, where 
$W_{\epsilon}'$ is an open set which contains $W_{\epsilon}$ and 
$g:\{u\in {\cal M};K(g(u))\neq 0\}\rightarrow \{w\in W_{\epsilon}';K(w)\neq 0\}$ is a bijective map, \\
(3) $a(x,u)$ is an $L^{s}(q)$-valued analytic function,  \\
(4) and $b(u)$ is an infinitely many times differentiable function which satisfies $b(u)>0$, \\
such that the following equations are satisfied in each local coordinate of ${\cal M}$. 
\begin{eqnarray*}
K(g(u))&=&u^{2k},\\
f(x,g(u))&=&u^{k}a(x,u),\\
\varphi(w)dw&=&\varphi(g(u))|g'(u)|du=
b(u)|u^{h}|du,
\end{eqnarray*}
where $k=(k_{1},k_{2},...,k_{d})$ and $h=(h_{1},h_{2},...,h_{d})$ 
are multi-indices made of nonnegative integers. At least one of $k_{j}$ is not equal to zero. 
\end{Theorem}
\vskip5mm
\noindent{\bf Remark}.
(1) In this theorem, for $u=(u_{1},u_{2},\cdots,u_{d})\in{\RR}^{d}$, 
notations $u^{2k}$ and $|u^{h}|$ 
respectively represent
\begin{eqnarray*}
u^{2k}&=&u_{1}^{2k_{1}}u_{2}^{2k_{2}}\cdots u_{d}^{2k_{d}},\\
|u^{h}|&=&|u_{1}^{h_{1}}u_{2}^{h_{2}}\cdots u_{d}^{h_{d}}|.
\end{eqnarray*}
The singularity $u=0$ in $u^{2k}=0$ is said to be normal crossing. Theorem \ref{Theorem:base}
shows that any singularities can be made normal crossing by using an analytic function $w=g(u)$.\\
(2) A map $w=g(u)$ is said to be proper if, for an arbitrary compact set $C$, 
$g^{-1}(C)$ is also compact. \\
(3) The proof of Theorem \ref{Theorem:base} is given in Theorem 6.1 of \citep{Cambridge2009} and
\citep{IWSMI2010}. 
In order to prove this theorem, we need the 
Hironaka resolution Theorem \citep{Hironaka,Atiyah}. 
The function $w=g(u)$ is often referred to as a resolution map. \\
(4) In this theorem, a quartet $(k,h, a(x,u),b(u))$ depends on 
a local coordinate in general. For a given function $K(w)$, there is an algebraic recursive algorithm 
which enables us to find a resolution map $w=g(u)$. However, for a fixed $K(w)$,  a resolution map
is not unique, resulting that a quartet $({\cal M}, g(u), a(x,u), b(u))$ is not unique. 
\vskip5mm\noindent
{\bf Definition}. (Real Log Canonical Threshold) 
Let $\{{\cal U}_{\alpha};\alpha\in{\cal A}\}$ be a system of local coordinates of a manifold ${\cal M}$,
\[
{\cal M}=\bigcup_{\alpha\in {\cal A}}{\cal U}_{\alpha}.
\]
The real log canonical threshold (RLCT) is defined by
\begin{equation}\label{eq:RLCT}
\lambda=\min_{\alpha\in{\cal A}}\min_{j=1}^{d}\Bigl(\frac{h_{j}+1}{2k_{j}}\Bigr),
\end{equation}
where we define $1/k_{j}=\infty$ for $k_{j}=0$. 
The multiplicity $m$ is defined by 
\begin{equation}\label{eq:mmm}
m=\max_{\alpha\in{\cal A}}\#\Bigl\{j;\frac{h_{j}+1}{2k_{j}}=\lambda\Bigr\},
\end{equation}
where $\# S$ shows the number of elements of a set $S$. 
\vskip5mm\noindent
This concept RLCT is well known in algebraic geometry and statistical learning theory. 
In the following definition we introduce a parity of a statistical model. 
\vskip5mm\noindent
{\bf Definition}. (Parity of Statistical Model) 
The support of $\varphi(g(u))$ is defined by 
\[
\mathrm{supp}\;\varphi(g(u))=\overline{\{u\in{\cal M}\;;\;
g(u)\in W_{\epsilon},\;\;\varphi(g(u))>0\}},
\]
where $\overline{S}$ shows the closure of a set $S$. 
A local coordinate ${\cal U}_{\alpha}$ is said to be an essential local coordinate
if both equations
\begin{eqnarray*}
\lambda&=&\min_{j=1}^{d}\Bigl(\frac{h_{j}+1}{2k_{j}}\Bigr),\\
m&=&\#\{j;(h_{j}+1)/(2k_{j})=\lambda\},
\end{eqnarray*}
hold  in its local coordinate. The set of all essential local coordinates is 
denoted by $\{{\cal U}_{\alpha};\alpha\in{\cal A}^{*}\}$. 
If, for an arbitrary essential local coordinate, there exist $\delta>0$ and 
a natural number $j$ in the set 
$
\{j\;;\;(h_{j}+1)/(2k_{j})=\lambda\}
$
such that \\
(1) $k_{j}$ is an odd number, \\
(2) $\{(0,0,..,0,u_{j},0,0,..,0)\;;\;|u_{j}|<\delta \}\subset \mathrm{supp}\;\varphi(g(u))$, \\
then we define $Q(K(g(u)),\varphi(g(u)))=1$. 
If otherwise, $Q(K(g(u)),\varphi(g(u)))=0$. 
If there exists a resolution map $w=g(u)$ such that $Q(K(g(u)),\varphi(g(u)))=1$, 
then we define
\begin{equation}\label{eq:QQQ}
Q(K(w),\varphi(w))=1.
\end{equation}
If otherwise $Q(K(w),\varphi(w))=0$. 
If $Q(K(w),\varphi(w))=1$, then the parity of a statistical model is said to be odd, otherwise even. 
\vskip5mm\noindent
It was proved in Theorem 2.4 of \citep{Cambridge2009}
that, for a given set $(q,p,\varphi)$, $\lambda$ and $m$ 
are independent of a choice of a resolution map. 
Such a value is called a birational invariant. 
The RLCT is a birational invariant. 

\begin{Lemma}\label{Lemma:QQQ}
If a true distribution $q(x)$ is realizable by a statistical model $p(x|w)$, then
the value 
$Q(K(g(u)), \varphi(g(u)))$ is independent of a choice of a resolution map $w=g(u)$. 
\end{Lemma}
Proof of this lemma is shown in Section \ref{section:proof}.
Lemma \ref{Lemma:QQQ} indicates that, if a 
true distribution is realizable by a statistical model, then
$Q(K(g(u)),\varphi(g(u)))$ is a birational invariant. 
The present paper proposes a conjecture that
 $Q(K(g(u)),\varphi(g(u)))$ is a birational invariant in general. 
By Lemma \ref{Lemma:QQQ}, this conjecture is proved if we can show the 
proposition that,  
for an arbitrary nonnegative analytic function $K(w)$, there 
exist $q(x)$ and $p(x|w)$ such that $K(w)$ is the Kullback-Leibler distance
from $q(x)$ to $p(x|w)$. 
\vskip5mm\noindent
{\bf Example.} Let $w=(a,b,c)\in{\RR}^{3}$ and 
\[
K(w)=(ab+c)^{2}+a^{2}b^{4},
\]
which is the Kullback-Leibler distance of a neural network model
in Example 1.6 of \citep{Cambridge2009}, where a true distribution is realizable by a
statistical model. 
The prior $\varphi(w)$ is defined by some nonzero function on a sufficiently large compact set. 
Let a system of local coordinates be 
\[
{\cal U}_{i}=\{(a_{i},b_{i},c_{i})\in{\RR}^{3}\}\;\;\;(i=1,2,3,4).
\]
A resolution map $g:{\cal U}_{1}\cup {\cal U}_{2}\cup {\cal U}_{3}\cup {\cal U}_{4} \rightarrow {\RR}^{3}$ in each
local coordinate is defined by 
\begin{eqnarray*}
a=a_{1}c_{1},\;\;& b= b_{1},\;\;&c=c_{1},\\
a=a_{2},\;\;&b=b_{2}c_{2},\;\;&c=a_{2}(1-b_{2})c_{2},\\
a=a_{3},\;\;&b=b_{3},\;\;&c=a_{3}b_{3}(b_{3}c_{3}-1),\\
a=a_{4},\;\;&b=b_{4}c_{4},\;\;&c=a_{4}b_{4}c_{4}(c_{4}-1).
\end{eqnarray*}
Then
\begin{eqnarray*}
K(a,b,c)&=& c_{1}^{2}\{(a_{1}b_{1}+1)^{2}+a_{1}^{2}b_{1}^{4}\}
=a_{2}^{2}c_{2}^{2}(1+b_{2}^{2}c_{2}^{2})\\
&=&a_{3}^{2}b_{3}^{4}(c_{3}^{2}+1)
=a_{4}^{2}b_{4}^{2}c_{4}^{4}(1+b_{4}^{2}).
\end{eqnarray*}
The Jacobian determinant $|g'(u)|$ is 
\begin{eqnarray*}
|g'(u)|&=&|c_{1}|
=|a_{2}c_{2}|\\
&=&|a_{3}b_{3}^{2}|
=|a_{4}b_{4}c_{4}|^{2}.
\end{eqnarray*}
Therefore $\lambda=3/4$ and $m=1$. The essential local coordinates are ${\cal U}_{3}$ and ${\cal U}_{4}$. 
In ${\cal U}_{3}$ and ${\cal U}_{4}$, the sets $\{u_{j}\;;\;(h_{j}+1)/(2k_{j})=3/4\}$ are respectively
$\{b_{3}\}$ and $\{c_{4}\}$, where $2k_{j}=4$ in both cases. Consequently, both $k_{j}$ are even, 
$Q(K(w),\varphi(w))=0$. 
\vskip5mm\noindent
\begin{Lemma}\label{Lemma:regular}
Assume that the Fundamental Conditions (1)-(4) are satisfied and 
that a true distribution $q(x)$ is regular for a statistical model $p(x|w)$. 
If $w_{0}$ is contained in the open kernel of $W$ and if $\varphi(w_{0})>0$, then 
\[
\lambda=\frac{d}{2},\;\;\;m=1,
\]
and
\[
Q(K(w),\varphi(w))=1.
\]
\end{Lemma}
Proof of this lemma is shown in Section \ref{section:proof}.

\begin{Theorem}\label{Theorem:NC2001}
Assume that the Fundamental Conditions (1)-(4) are satisfied. Then the 
following holds. 
\[
{\cal F}=nL_{n}(w_{0})+\lambda\log n- (m-1)\log\log n + R_{n},
\]
where $\lambda$ is a real log canonical threshold, $m$ is its multiplicity, and 
$R_{n}$ is an random variable which converges to a random variable in law, 
when $n\rightarrow\infty$. 
\end{Theorem}
Theorem \ref{Theorem:NC2001} was proved in the previous papers. 
In the case when $q(x)$ is realizable by and singular for
$p(x|w)$, the  expectation value of ${\cal F}$ is given by \citep{NC2001}.
The asymptotic behavior of ${\cal F}$ as a random variable was shown in 
\citep{Cambridge2009}.
These results were generalized in \citep{IWSMI2010} for the case that $q(x)$ is unrealizable. 
\vskip5mm\noindent
{\bf Remark.} In practical applications, we do not know the true distribution, 
hence $\lambda$ and $m$ are unknown. Therefore, we can not directly apply Theorem \ref{Theorem:NC2001}
to such cases. The main purpose of the present paper is to make a new method how to estimate 
${\cal F}$ even if the true distribution is unknown.

\section{Main Results}

In this section, we introduce the main results of the present paper. 

\begin{Theorem}$($Unique Existence of the Optimal Parameter$)$ 
\label{Theorem:unique}
Assume that $L_{n}(w)$ is not a constant function of $w$. Then 
the followings hold.\\
(1) The value $\EE_{w}^{\beta}[nL_{n}(w)]$ is a decreasing function
of $\beta$. \\
(2) There exists a unique $\beta^{*}$ $(0<\beta^{*}<1)$ which satisfies
\begin{equation}\label{eq:unique}
{\cal F}=\EE_{w}^{\beta^{*}}[nL_{n}(w)].
\end{equation}
\end{Theorem}
The Proof of Theorem \ref{Theorem:unique} is given in Section \ref{section:proof}. 
Based on this theorem, we define the optimal inverse 
temperature. 
\vskip5mm\noindent
{\bf Definition}. The unique parameter $\beta^{*}$ that satisfies
eq.(\ref{eq:unique}) is called the optimal inverse temperature. 
\vskip5mm\noindent
In general, the optimal inverse temperature $\beta^{*}$ depends on
a true distribution $q(x)$, a statistical model $p(x|w)$, a prior $\varphi(w)$, and
training samples. Therefore $\beta^{*}$ is a random variable. In the present paper,
we study its probabilistic behavior. Theorem \ref{Theorem:main} is a mathematical base 
for such a purpose. 

\begin{Theorem}$($Main Theorem$)$ \label{Theorem:main}
Assume Fundamental Conditions (1)-(4) and  that
\[
\beta=\frac{\beta_{0}}{\log n},
\]
where $\beta_{0}$ is a constant. 
Then there exists a random variable $U_{n}$ such that
\[
\EE_{w}^{\beta}[nL_{n}(w)]=nL_{n}(w_{0})
+\frac{\lambda\log n}{\beta_{0}} + U_{n}
\sqrt{\frac{\lambda\log n}{2\beta_{0}}}+O_{p}(1),
\]
where $\lambda$ is the real log canonical threshold and 
$U_{n}$ is a random variable, which satisfies 
$
\EE[U_{n}]=0,
$ 
converges to a gaussian random variable in law as $n\rightarrow \infty$. 
Moreover, if a true distribution $q(x)$ is realizable by a statistical model $p(x|w)$, then 
$
\EE[(U_{n})^2]< 1.
$
\end{Theorem}
\vskip5mm\noindent
The proof of Theorem \ref{Theorem:main} is given in Section \ref{section:proof}.
Theorem \ref{Theorem:main} with $\beta_{0}=1$ shows that 
\[
\mathrm{WBIC}=nL_{n}(w_{0})
+\lambda\log n + U_{n}
\sqrt{\frac{\lambda\log n}{2}}+O_{p}(1),
\]
whose first two main terms are equal to those of ${\cal F}$ in Theorem \ref{Theorem:NC2001}. 
From Theorem \ref{Theorem:main}
and its proof, three important corollaries are derived. 

\begin{Corollary} \label{Corollary:111}
If the parity of a statistical model is odd, $Q(K(w),\varphi(w))=1$, then $U_{n}=0$. 
\end{Corollary}

\begin{Corollary} \label{Corollary:222} 
Let $\beta^{*}$ be the optimal inverse temperature. Then
\[
\beta^{*}=\frac{1}{\log n}\Bigl(1+\frac{U_{n}}{\sqrt{2\lambda\log n}}
+o_{p}\Bigl(\frac{1}{\sqrt{\log n}}\Bigr)\Bigr).
\]
\end{Corollary}

\begin{Corollary}\label{Corollary:333} 
Let $\beta_{1}=\beta_{01}/\log n$ and  $\beta_{2}=\beta_{02}/\log n$,
where $\beta_{01}$ and $\beta_{02}$ are positive constants.  
Then the convergence in probability
\begin{equation}\label{eq:lambda-estimated}
\frac{
\EE_{w}^{\beta_{1}}[nL_{n}(w)]-\EE_{w}^{\beta_{2}}[nL_{n}(w)]
}{
1/\beta_{1}-1/\beta_{2}
}
\rightarrow \lambda
\end{equation}
holds as $n\rightarrow \infty$, where $\lambda$ is the real log canonical threshold. 
\end{Corollary}
Proofs of these corollaries are given in Section \ref{section:proof}.
\vskip5mm\noindent

The well-known Schwarz BIC is defined by
\[
\mathrm{BIC}=nL_{n}(\hat{w})+\frac{d}{2}\log n,
\]
where $\hat{w}$ is the maximum likelihood estimator. 
WBIC can be understood as the generalized BIC onto singular
statistical models, because it satisfies the following theorem. 

\begin{Theorem}\label{Theorem:regular}
If a true distribution $q(x)$ is regular for 
a statistical model $p(x|w)$, then 
\[
\mathrm{WBIC}=nL_{n}(\hat{w})+\frac{d}{2}\log n+o_{p}(1).
\]
\end{Theorem} 
Proof of Theorem \ref{Theorem:regular} is given in Section \ref{section:proof}.
This theorem shows that the difference of 
WBIC and BIC is smaller than a constant order term, if
a true distribution is regular for a statistical model. 
This theorem holds even if a true distribution
$q(x)$ is unrealizable by $p(x|w)$. 
\vskip5mm\noindent
{\bf Remark.} Since the set of parameters $W$ is assumed to be compact, 
it is proved in Main Theorem 6.4 of \citep{Cambridge2009} that 
$nL_{n}(w_{0})-nL_{n}(\hat{w})$ is a constant order random variable in general. 
If a true distribution is regular for and realizable by a statistical model,
its average is asymptotically equal to $d/2$, where $d$ is the dimension of parameter. 
However, if  
a true distribution is singular for a statistical model, then it is much larger than $d/2$,
because it is asymptotically equal to the maximum value of the Gaussian process. Hence
replacement of $nL_{n}(w_{0})$ by $nL(\hat{w})$ is not appropriate
in singular model evaluation.

\section{Proofs of Main Results}\label{section:proof}

In this section, we prove the main theorems and corollaries. 

\subsection{Proof of Lemma \ref{Lemma:start}}

Let us define an empirical process, 
\[
\eta_{n}(w)=\frac{1}{\sqrt{n}}\sum_{i=1}^{n}(K(w)-f(X_{i},w)).
\]
It was proved  in Theorem 5.9 and 5.10 of \citep{Cambridge2009} that
$\eta_{n}(w)$ converges to a random process in law and 
\[
\|\eta_{n}\|\equiv \sup_{w\in W}|\eta_{n}(w)|
\]
also converges to a random variable in law.  
If $K(w)\geq 1/n^{r}$, then 
\begin{eqnarray*}
nK_{n}(w)&=&nK(w)-\sqrt{n}\;\eta_{n}(w) \\
&\geq & n^{1-r} -\sqrt{n}\;\|\eta_{n}\|. 
\end{eqnarray*}
By the condition $1-r>1/2$ and $\beta=\beta_{0}/\log n$, 
\begin{eqnarray*}
\exp(\sqrt{n})\int_{K(w)\geq 1/n^{r}}\exp(-n\beta K_{n}(w))\varphi(w)dw \\
\leq \exp(-n^{1-r}\beta+\sqrt{n}+\sqrt{n}\beta \|\eta_{n}\|),
\end{eqnarray*}
which converges to zero in probability, which shows eq.(\ref{eq:r11}). 
Then, let us prove eq.(\ref{eq:r12}). 
Since the set of parameter $W$ is compact, $\|K\|\equiv \sup_{w} K(w)<\infty$. Therefore,
\begin{eqnarray*}
|nK_{n}(w)|&\leq & n\|K\|+\sqrt{n}\|\eta_{n}\| \\
&= & n\;(\|K\|+\|\eta_{n}\|/\sqrt{n}).
\end{eqnarray*}
Hence 
\begin{eqnarray*}
&&\exp(\sqrt{n})\int_{K(w)\geq 1/n^{r}}|nK_{n}(w)|\exp(-n\beta K_{n}(w))\varphi(w)dw \\
&&\leq (\|K\|+\|\eta_{n}\|/\sqrt{n})\\
&&\times \exp(-n^{1-r}\beta  +\sqrt{n}+\sqrt{n}\beta \|\eta_{n}\|+\log n),
\end{eqnarray*}
which converges to zero in probability. (Q.E.D.) 

\subsection{Proof of Lemma \ref{Lemma:regular}}

Without loss of generality, we can assume $w_{0}=0$. 
Since $q(x)$ is regular for $p(x|w)$, there exists $w^{*}$ such that
\[
K(w)=\frac{1}{2}w\cdot J(w^{*})w,
\]
where $J(w)$ is given in eq.(\ref{eq:Jij}). Since $J(w_{0})$ is a strictly positive
definite matrix, there exists $\epsilon>0$ such that, if  $K(w)\leq \epsilon$, then $J(w^{*})$ 
is positive definite. Let $\ell_{1}$ and $\ell_{2}$ be respectively the minimum and maximum 
eigen values of $\{J(w^{*});K(w)\leq \epsilon\}$. Then
\[
\frac{1}{4}\ell_{1}\sum_{j=1}^{d}w_{j}^{2}\leq \frac{1}{2}w\cdot J(w^{*})w\leq \ell_{2}\sum_{j=1}^{d}w_{j}^{2}.
\]
By using a blow-up $g:{\cal U}_{1}\cup \cdots \cup {\cal U}_{d} \rightarrow W$ 
which is represented on each local coordinate ${\cal U}_{i}=(u_{i1},u_{i2},...,u_{id})$, 
\begin{eqnarray*}
w_{i}&=&u_{ii},\\
w_{j}&=&u_{ii}u_{ij}\;\;\;(j\neq i),
\end{eqnarray*}
it follows that
\[
\frac{\ell_{1}\;u_{ii}^{2}}{4}(1+\sum_{j\neq i}u_{ij}^{2})
\leq \frac{u_{ii}^{2}}{2}(\hat{u},J(w^{*})\hat{u})
\leq \ell_{2}\;u_{ii}^{2}(1+\sum_{j\neq i}u_{ij}^{2}),
\]
where $\hat{u}_{ij}=u_{ij}$ $(j\neq i)$ and $\hat{u}_{ii}=1$. 
These inequalities show that $k_{i}=1$ in ${\cal U}_{i}$,  
therefore $Q(K(w),\varphi(w))=1$.
The Jacobian determinant of the blow-up is 
\[
|g'(u)|=|u_{ii}|^{d-1},
\]
hence $\lambda=d/2$ and $m=1$.  (Q.E.D.)

\subsection{Proof of Theorem \ref{Theorem:unique}}

Let us define a function $F_{n}(\beta)$ of $\beta>0$ by
\[
F_{n}(\beta)=-\log \int \prod_{i=1}^{n}
p(X_{i}|w)^{\beta}\varphi(w)dw. 
\]
Then, by the definition, ${\cal F}=F_{n}(1)$ and 
\begin{eqnarray*}
F_{n}'(\beta)&=&\EE_{n}^{\beta}[nL_{n}(w)], \\
F_{n}''(\beta)&=&-\EE_{n}^{\beta}[(nL_{n}(w))^{2}]+\EE_{n}^{\beta}[nL_{n}(w)]^{2}. 
\end{eqnarray*}
By the Cauchy-Schwarz inequality and the assumption that $L_{n}(w)$ is not a 
constant function, 
\[
F_{n}''(\beta)<0,
\]
which shows (1). Since $F_{n}(0)=0$, 
\[
{\cal F}=F_{n}(1)=\int_{0}^{1}F_{n}'(\beta)d\beta.
\]
By using the mean value theorem, there exists $\beta^{*}$ 
($0<\beta^{*}<1$) such that 
\[
{\cal F}=F_{n}'(\beta^{*})=\EE_{n}^{\beta^{*}}[nL_{n}(w)].
\]
Here $F_{n}'(\beta)$ is a decreasing function, $\beta^{*}$ is
unique, which completes Theorem \ref{Theorem:unique}. (Q.E.D.)

\subsection{First Preparation for Proof of Theorem \ref{Theorem:main}}

In this subsection, we prepare the proof of Theorem  \ref{Theorem:main}. 
By using eq.(\ref{eq:LnKn}) and eq.(\ref{eq:KnKn}), the proof of Theorem \ref{Theorem:main} 
results in evaluating $E_{w}^{\beta}[nK_{n}(w)]$.  
By Lemma \ref{Lemma:start},
\begin{equation}\label{eq:ZZZ}
E_{w}^{\beta}[nK_{n}(w)]= \frac{B_{n}+o_{p}(\exp(-\sqrt{n}))}{A_{n}+o_{p}(\exp(-\sqrt{n}))},
\end{equation}
where $A_{n}$ and $B_{n}$ are respectively defined by 
\begin{eqnarray}
A_{n}&=& \int_{K(w)<\epsilon}\exp(-n\beta K_{n}(w))\varphi(w)dw, \label{eq:An} \\
B_{n}&=& \int_{K(w)<\epsilon}nK_{n}(w)\exp(-n\beta K_{n}(w))\varphi(w)dw. \label{eq:Bn}
\end{eqnarray}
By Theorem \ref{Theorem:base}, an integral over $\{w\in W;K(w)<\epsilon\}$ can be calculated 
by that over ${\cal M}$. 
For a given local coordinates $\{{\cal U}_{\alpha}\}$ of ${\cal M}$,
there exists a set of $C^{\infty}$ class functions $\{\varphi_{\alpha}(g(u))\}$ 
such that, for an arbitrary $u\in{\cal M}$, 
\[
\sum_{\alpha\in {\cal A}}\varphi_{\alpha}(g(u))=\varphi(g(u)). 
\]
By using this fact, for arbitrary integrable function $G(w)$, 
\[
\int_{K(w)<\epsilon} G(w)\varphi(w) dw = 
\sum_{\alpha\in {\cal A}} \int_{\cal {\cal U}_{\alpha}}G(g(u))\varphi_{\alpha}(g(u))|g'(u)|du.
\]
Without loss of generality, we can assume that 
$\overline{{\cal U}_{\alpha}\cap \mathrm{supp}\;\varphi(g(u))}$ 
is isomorphic to $[-1,1]^{d}$
and that $\varphi_{\alpha}(g(u))>0$ in $[-1,1]^{d}$. 
Moreover, by Theorem \ref{Theorem:base}, there exists a function $b_{\alpha}(u)>0$ such that 
\[
\varphi_{\alpha}(g(u))|g'(u)|=|u^{h}|b_{\alpha}(u),
\]
in each local coordinate. Consequently, 
\[
\int_{K(w)<\epsilon} G(w)\varphi(w) dw = \sum_{\alpha\in{\cal A}} \int_{[-1,1]^{d}}du\;G(g(u))\;|u^{h}|\;b_{\alpha}(u).
\]
In each local coordinate, 
\[
K(g(u))=u^{2k}.
\]
We define a function $\xi_{n}(u)$ by 
\[
\xi_{n}(u)=\frac{1}{\sqrt{n}}\sum_{i=1}^{n}\{u^{k}-a(X_{i},u)\}.
\]
Then 
\[
K_{n}(g(u))=u^{2k}-\frac{1}{\sqrt{n}}u^{k}\xi_{n}(u).
\]
Note that
\[
u^{k}=\int a(x,u)q(x)dx
\]
holds, because 
\[
u^{2k}=\int f(x,g(u))q(x)dx=u^{k}\int a(x,u)q(x)dx. 
\]
Therefore, for an arbitrary $u$, 
\[
\EE[\xi_{n}(u)]=0.
\]
The function $\xi_{n}(u)$ can be understood as a random process on ${\cal M}$. 
On Fundamental Conditions (1)-(4), it is proved in Theorem 6.1, Theorem 6.2, and Theorem 6.3 
of \citep{Cambridge2009}
that \\
(1) $\xi_{n}(u)$ converges to a gaussian random process $\xi(u)$ in law and  
\[
\EE[\sup_{u}\xi_{n}(u)^{2}]\rightarrow\EE[\sup_{u}\xi(u)^{2}]. 
\]
(2) If $q(x)$ is realizable by $p(x|w)$, and if $u^{2k}=0$, then 
\begin{equation}\label{eq:xi(u)=2}
\EE[\xi_{n}(u)^{2}]=\EE_{X}[a(X,u)^2]=2.
\end{equation}
By using the random process $\xi_{n}(u)$,  
two random variable $A_{n}$ and $B_{n}$ can be represented by integrals over ${\cal M}$, 
\begin{eqnarray}
A_{n}&=& \sum_{\alpha\in{\cal A}}\int_{[-1,1]^{d}}du\;
\exp(-n\beta u^{2k}+\sqrt{n}\beta u^{k} \xi_{n}(u))|u^{h}|b_{\alpha}(u), \label{eq:An2} \\
B_{n}&=& \sum_{\alpha\in{\cal A}}\int_{[-1,1]^{d}}du\;(n u^{2k}-\sqrt{n}u^{k}\xi_{n}(u))\nonumber \\
&& \times 
\exp(-n\beta u^{2k}+\sqrt{n}\beta u^{k}\xi_{n}(u))|u^{h}|b_{\alpha}(u) \label{eq:Bn2}.
\end{eqnarray}
To prove Theorem \ref{Theorem:main}, we study asymptotics of these two values. 

\subsection{Second Preparation for Proof of Theorem \ref{Theorem:main}}

To evaluate two integrals $A_{n}$ and $B_{n}$ as $n\rightarrow\infty$, 
we have to study the asymptotic behavior of the following Schwartz distribution,
\[
\delta(t-u^{2k})\;|u|^{h}
\]
for $t\rightarrow 0$. 
Without loss of generality, 
we can assume that, in each essential local coordinate, 
\[
\lambda=
\frac{h_{1}+1}{2k_{1}}=
\frac{h_{2}+1}{2k_{2}}=\cdots=
\frac{h_{m}+1}{2k_{m}}<\frac{h_{j}+1}{2k_{j}},
\]
where $m<j\leq d$. A variable $u\in {\RR}^{d}$ is denoted by 
\[
u=(u_{a},u_{b})\in {\RR}^{m}\times {\RR}^{d-m}.
\]
We define a measure $du^{*}$ by 
\begin{equation}\label{eq:u*}
du^{*}=\frac{\displaystyle( \prod_{j=1}^{m}\delta(u_{j}))\;(
\prod_{j=m+1}^{d}(u_{j})^{\mu_{j}})\;du}
{2^{m}\;(m-1)!\;(\prod_{j=1}^{m}k_{j})},
\end{equation}
where  $\delta(\;\;)$ is the Dirac delta function, 
and $\mu=(\mu_{m+1},\mu_{2},...,\mu_{d})$ is a multi-index defined by
\[
\mu_{j}=-2\lambda k_{j}+h_{j}\;\;\;(m+1\leq j\leq d).
\]
Then $\mu_{j}>-1$, hence eq.(\ref{eq:u*}) defines
a measure on ${\cal M}$. The support of $du^{*}$ is 
$\{u=(u_{a},u_{b})\;;\;u_{a}=0\}$. 
\vskip5mm\noindent
{\bf Definition}. Let $\sigma$ be a $d$dimensional 
variable, 
\[
\sigma=(\sigma_{1},\sigma_{2},...,\sigma_{d})\in {\RR}^{d}
\]
where $\sigma_{j}=\pm 1$. The set of all such 
variables is denoted by $S(d)$. We use a notation
\[
\sigma u=(\sigma_{1}u_{1},\sigma_{2}u_{2},...,\sigma_{d}u_{d})\in{\RR}^{d}.
\]
Then $(\sigma u)^{k}=\sigma^{k}u^{k}$ and  $(\sigma u)^{2k}=u^{2k}$.
By using this notation, we
can derive the asymptotic behavior of $\delta(t-u^{2k})|u^{h}|$ for $t\rightarrow 0$. 

\begin{Lemma} \label{Lemma:singulardist}
Let $G(u^{2k},u^{k},u)$ be a real-valued $C_{1}$-class function of $(u^{2k},u^{k},u)$
($u\in{\RR}^{d}$). 
The following asymptotic expansion holds as $t\rightarrow +0$, 
\begin{eqnarray}
&&\int_{[-1,1]^{d}}du\;\delta(t-u^{2k})
|u|^{h}G(u^{2k},u^{k},u)\nonumber \\
&&=t^{\lambda-1}(-\log t)^{m-1}\sum_{\sigma\in S(d)}
\int_{[0,1]^{d}}du^{*}\;
G(t,\sigma^{k}\sqrt{t},u)\nonumber \\
&&+O\Bigl(t^{\lambda-1}(-\log t)^{m-2}\Bigr), \label{eq:delta}
\end{eqnarray}
where $du^{*}$ is a measure defined by eq.(\ref{eq:u*}). 
\end{Lemma}
(Proof of Lemma \ref{Lemma:singulardist}) Let $Y(t)$ 
be the left hand side of eq.(\ref{eq:delta}).
\begin{eqnarray*}
Y(t)
&=& \sum_{\sigma\in S(d)}
\int_{[0,1]^{d}}\delta(t-(\sigma u)^{2k})
|\sigma u|^{h}G((\sigma u)^{2k},(\sigma u)^{k},\sigma u)d(\sigma u)\\
&=& \sum_{\sigma\in S(d)}
\int_{[0,1]^{d}}\delta(t-u^{2k})
|u|^{h}G(t,\sigma^{k}\sqrt{t},u)du.
\end{eqnarray*}
By using Theorem 4.9 of \citep{Cambridge2009}, if $u\in [0,1]^{d}$, then
\begin{eqnarray*}
\delta(t-u^{2k})|u|^{h}du&=& t^{\lambda-1}(-\log t)^{m-1}du^{*} \\
&&+O(t^{\lambda-1}(-\log t)^{m-1}).
\end{eqnarray*}
By applying this relation to $Y(t)$, 
we obtain Lemma \ref{Lemma:singulardist}. (Q.E.D.) 

\subsection{Proof of Lemma \ref{Lemma:QQQ}}

Let $\Phi(w)>0$ be an arbitrary $C^{\infty}$ class function
on $W_{\epsilon}$. 
Let $Y(t,\Phi)$ ($t>0$) be a function defined by 
\[
Y(t,\Phi)\equiv \int_{K(w)<\epsilon}\delta(t-K(w))f(x,w)\Phi(w)\varphi(w)dw,
\]
whose value is independent of 
a choice of a resolution map. 
By using a resolution map $w=g(u)$, 
\[
Y(t,\Phi)=\sum_{\alpha\in{\cal A}}\sum_{\sigma\in S(d)}
\int_{[-1,1]^{d}} du\;\delta(t-u^{2k})\;u^{k}\;|u|^{h}a(x,u)\Phi(g(u))b_{\alpha}(u)du.
\]
By Lemma \ref{Lemma:singulardist}, and $\sigma=(\sigma_{a},\sigma_{b})$, 
\begin{eqnarray*}
Y(t,\Phi)&=&t^{\lambda-1/2}(\log t)^{m-1}
\sum_{\alpha\in {\cal A}^{*}}\sum_{\sigma_{a}\in S(m)}(\sigma_{a})^{k}\sum_{\sigma_{b}\in S(d-m)}(\sigma_{b})^{k}
\\
&& \times
\int_{[0,1]^{d}} du^{*}\;a(x,\sigma u)\;\Phi(g(\sigma u))\;b_{\alpha}(\sigma u)\\
&&+ O(t^{\lambda-1/2}(\log t)^{m-2}). 
\end{eqnarray*}
By the assumption that a
true distribution is realizable by a statistical model, eq.(\ref{eq:xi(u)=2}) shows that 
there exists $x$ such that $a(x,u)\neq 0$ for $u^{2k}=0$. 
On the support of $du^{*}$,  
\[
\sigma u= (\sigma_{a}u_{a},\sigma_{b}u_{b})=(0,\sigma_{b}u_{b}),
\]
consequently the main order term of $Y(t,\Phi(w))$ is determined by $\Phi(0,u_{b})$. 
If $Q(K(g(u)), \varphi(g(u)))=1$, then at least one $k_{j}$ $(1\leq j\leq m)$ is 
odd, $\sigma_{a}^{k}$ takes values $\pm 1$, hence 
\[
\sum_{\sigma_{a} \in S(m)}\sigma_{a}^{k}=0,
\]
which shows that the coefficient 
of the main order term in $Y(t,\Phi)$ $(t\rightarrow +0)$ is zero for
an arbitrary $\Phi(w)$. If  $Q(K(g(u)),\varphi(g(u)))=0$,
\[
\sum_{\sigma_{a} \in S(m)}\sigma_{a}^{k}\sum_{\sigma_{a} \in S(m)}1\neq 0.
\]
There exists a function $\Phi(w)$ such that them main order term is not equal to zero.
Therefore $Q(K(g(u)),\varphi(g(u)))$ does not depend on the resolution map. 
 (Q.E.D.)

\subsection{Proof of Theorem \ref{Theorem:main}}

In this subsection, we prove Theorem \ref{Theorem:main} using the foregoing preparations.
We need to study $A_{n}$ and $B_{n}$ in eq.(\ref{eq:An}) and eq.(\ref{eq:Bn}). 
Firstly, we study $A_{n}$.
\begin{eqnarray*}
A_{n}
&=&\sum_{\alpha\in{\cal A}}\int_{[-1,1]^{d}}du\; 
\exp(-n\beta u^{2k}+\beta \sqrt{n}u^{k}\xi_{n}(u))|u|^{h}b_{\alpha}(u) \\
&=&\sum_{\alpha\in{\cal A}}\int_{[-1,1]^{d}}du \int_{0}^{\infty} dt 
\;\delta(t-u^{2k})|u|^{h}b_{\alpha}(u)\\
&&\times \exp(-n\beta u^{2k}+\beta \sqrt{n}u^{k}\xi_{n}(u)).
\end{eqnarray*}
By substitution $t:=t/(n \beta)$ and $dt:=dt/(n\beta)$, 
\begin{eqnarray*}
A_{n}&=&\sum_{\alpha\in{\cal A}}\int_{[-1,1]^{d}} b_{\alpha}(u)du \int_{0}^{\infty} 
\frac{dt}{n\beta} \delta\Bigl(\frac{t}{n \beta } -u^{2k}\Bigr)|u|^{h}\\
&&\times \exp(-n\beta u^{2k}+\beta \sqrt{n}u^{k}\xi_{n}(u)).
\end{eqnarray*}
For simple notations, we use 
\begin{eqnarray*}
\int_{\cal M} du^{*}&\equiv &\sum_{\alpha\in{\cal A}^{*}}
\sum_{\sigma\in S(d)}\int_{[0,1]^{d}}b_{\alpha}(u)\;du^{*},\\
\xi_{n}^{*}(u)&\equiv &\sigma^{k}\xi_{n}(u), 
\end{eqnarray*}
where $\{{\cal U}_{\alpha}\;;\;\alpha\in{\cal A}^{*}\}$ is the set of all 
essential local coordinates. 
Then by using Lemma \ref{Lemma:singulardist}, $\delta(t/n\beta-u^{2k})$ can be 
asymptotically expanded for $n\beta\rightarrow 0$, hence 
\begin{eqnarray*}
A_{n}&=&\int_{\cal M}du^{*}\int_{0}^{\infty} \frac{dt}{n\beta}
\;\Bigl(\frac{t}{n\beta}\Bigr)^{\lambda-1} \Bigl(-\log (\frac{t}{n\beta})\Bigr)^{m-1} \\
&&\times \exp(-t+\sqrt{\beta t}\;\xi^{*}_{n}(u)) + O_{p}(\frac{(\log (n\beta))^{m-2}}{(n\beta)^{\lambda}}) \\
&=& 
\frac{(\log (n\beta))^{m-1}}{(n\beta)^{\lambda}}\int_{\cal M}du^{*}
 \int_{0}^{\infty} dt\;t^{\lambda-1} 
\exp(-t)\;\exp(\sqrt{\beta t}\;\xi_{n}^{*}(u)) \\
&& +O_{p}(\frac{(\log (n\beta))^{m-2}}{(n\beta)^{\lambda}}) .
\end{eqnarray*}
Since $\beta=\beta_{0}/\log n \rightarrow  0$, 
\[
\exp(\sqrt{\beta t}\;\xi_{n}^{*}(u))=
1+\sqrt{\beta t}\;\xi_{n}^{*}(u)+O_{p}(\beta).
\]
By using the gamma function,
\[
\Gamma(\lambda)=\int_{0}^{\infty}t^{\lambda-1}\;\exp(-t)\;dt, 
\]
it follows that 
\begin{eqnarray*}
A_{n}&=&
\frac{(\log (n\beta))^{m-1}}{(n\beta)^{\lambda}} \Bigl\{ \Gamma(\lambda)
\Bigl(\int_{\cal M} du^{*}\Bigr)
+ \sqrt{\beta}\Gamma(\lambda+\frac{1}{2})\Bigl(\int_{\cal M}  du^{*}\xi_{n}^{*}(u) \Bigr)\Bigr\} \\
&& +O_{p}(\frac{(\log (n\beta))^{m-2}}{(n\beta)^{\lambda}}).
\end{eqnarray*}
Secondly, $B_{n}$ can be calculated by the same way, 
\begin{eqnarray*}
B_{n}&=&\sum_{\alpha\in{\cal A}}\int_{[-1,1]^{d}}du \int_{0}^{\infty} dt \;\delta(t-u^{2k})|u|^{h}b_{\alpha}(u)\\
&&\times (n u^{2k}-\sqrt{n}u^{k}\xi_{n}(u))\exp(-n\beta u^{2k}+\beta \sqrt{n}u^{k}\xi_{n}(u)).
\end{eqnarray*}
By substitution $t:=t/(n \beta)$ and $dt:=dt/(n\beta)$ and Lemma \ref{Lemma:singulardist}, 
\begin{eqnarray*}
B_{n}&=&\int_{\cal M}du^{*}
 \int_{0}^{\infty} \frac{dt}{n\beta}\;\Bigl(\frac{t}{n\beta}\Bigr)^{\lambda-1} 
\Bigl(-\log (\frac{t}{n\beta})\Bigr)^{m-1} \\
&&\times \frac{1}{\beta}(t-\sqrt{\beta t}\;\xi_{n}^{*}(u))
\exp(-t+\sqrt{\beta t}\;\xi_{n}^{*}(u)) + O_{p}(\frac{(\log (n\beta))^{m-2}}{\beta(n\beta)^{\lambda}}) \\
&=& 
\frac{(\log (n\beta))^{m-1}}{\beta (n\beta)^{\lambda}}\int_{\cal M}du^{*}
\int_{0}^{\infty} t^{\lambda-1}(t-\sqrt{\beta t}\;\xi_{n}^{*}(u))
\exp(-t) \\
&&\times \exp(\sqrt{\beta t}\;\xi_{n}^{*}(u)) +O_{p}(\frac{(\log (n\beta))^{m-2}}{\beta(n\beta)^{\lambda}}).
\end{eqnarray*}
Therefore, 
\begin{eqnarray*}
B_{n}&=&
\frac{(\log (n\beta))^{m-1}}{ \beta (n\beta)^{\lambda}} \Bigl\{ \Gamma(\lambda+1)
\Bigl(\int_{\cal M} du^{*}\Bigr)
+ \sqrt{\beta}\;\Gamma(\lambda+\frac{3}{2})\Bigl(\int_{\cal M} du^{*}\xi_{n}^{*}(u)\Bigr)\\
&& - \sqrt{\beta}\;\Gamma(\lambda+\frac{1}{2})\Bigl(\int_{\cal M} du^{*}\xi_{n}^{*}(u)\Bigr)\Bigr\} 
+O_{p}(\frac{(\log (n\beta))^{m-2}}{\beta(n\beta)^{\lambda}}).
\end{eqnarray*}
Let 
\begin{equation}\label{eq:def_Th}
\Theta=\frac{\int_{\cal M} du^{*}\xi_{n}^{*}(u)}{
\int_{\cal M} du^{*}}.
\end{equation}
By applying results of $A_{n}$ and $B_{n}$ to eq.(\ref{eq:ZZZ}), 
\[
\EE_{w}^{\beta}[nK_{n}(w)]=
\frac{1}{\beta}\times
\frac{\Gamma(\lambda+1)+\sqrt{\beta}\;\Theta\;\{\Gamma(\lambda+3/2)-\Gamma(\lambda+1/2)\}}
{\Gamma(\lambda)+\sqrt{\beta}\;\Theta\Gamma(\lambda+1/2)}+O_{p}(1).
\]
Note that, if $a,b,c,d$ are constants and $\beta\rightarrow 0$,
\[
\frac{c+\sqrt{\beta}\;d}{a+\sqrt{\beta}\;b}
=\frac{c}{a}+\sqrt{\beta}\;\Bigl(\frac{ad-bc}{a^{2}}\Bigr)+O(\beta). 
\]
Then by using an identity, 
\[
\frac{\Gamma(\lambda)(\Gamma(\lambda+3/2)-\Gamma(\lambda+1/2))
-\Gamma(\lambda+1)\Gamma(\lambda+1/2)}
{\Gamma(\lambda)^{2}}
=-\frac{\Gamma(\lambda+1/2)}{2\Gamma(\lambda)},
\]
we obtain 
\begin{eqnarray*}
&&\EE_{w}^{\beta}[nK_{n}(w)]=
\frac{1}{\beta}
\frac{\Gamma(\lambda+1)}{\Gamma(\lambda)} 
-\frac{\Theta}{\sqrt{\beta}} \frac{\Gamma(\lambda+1/2)}{2\Gamma(\lambda)}
+O_{p}(1). 
\end{eqnarray*}
A random variable $U_{n}$ is defined by  
\begin{equation}\label{eq:Un}
U_{n}=-\frac{\Theta \Gamma(\lambda+1/2)}{\sqrt{2\lambda}\Gamma(\lambda)}.
\end{equation}
Then it follows that 
\begin{eqnarray*}
&& \EE_{w}^{\beta}[nK_{n}(w)]=
\frac{\lambda}{\beta}
+U_{n}\sqrt{\frac{\lambda}{2\beta}}
+O_{p}(1).
\end{eqnarray*}
By the definition of $\xi_{n}(u)$,
$
\EE[\Theta]=0,$ hence $\EE[U_{n}]=0$.
By using Cauchy-Schwarz inequality, 
\[
\Theta^{2}\leq \frac{\int_{\cal M} du^{*}\;\xi_{n}^{*}(u)^{2}}{
\int_{\cal M} du^{*}}.
\]
Lastly let us study the case that $q(x)$ is realizable by $p(x|w)$.
The support of $du^{*}$ is contained in $u^{2k}=0$, hence we can apply eq.(\ref{eq:xi(u)=2}) 
to $\Theta$, 
\[
\EE[\Theta^{2}]\leq \frac{\int_{\cal M} du^{*}\EE[\xi_{n}^{*}(u)^{2}]}{
\int_{\cal M} du^{*}}=2. 
\]
The gamma function satisfies 
\[
\frac{\Gamma(\lambda+1/2)}{\Gamma(\lambda)}< \sqrt{\lambda}\;\;\;(\lambda>0). 
\]
Hence we obtain 
\[
\EE[(U_{n})^{2}]\leq \frac{\EE[\Theta^{2}]}{2\lambda}
\Bigl(\frac{\Gamma(\lambda+1/2)}{\Gamma(\lambda)}\Bigr)^{2}
< 1,
\]
which completes Theorem \ref{Theorem:main}. 
(Q.E.D.)

\subsection{Proof of Corollary \ref{Corollary:111}}

By definition eq.(\ref{eq:def_Th}) and eq.(\ref{eq:Un}), it is sufficient to 
prove $\Theta=0$, where 
\[
\Theta=
\frac{\displaystyle
\sum_{\alpha\in{\cal A}^{*}}\sum_{\sigma\in S(d)}
\int_{[0,1]^{d}} b_{\alpha}(u)\;du^{*}\;\sigma^{k}\;
\xi_{n}(u)
}{\displaystyle
\sum_{\alpha\in{\cal A}^{*}}\sum_{\sigma\in S(d)}\int_{[0,1]^{d}} b_{\alpha}(u)\;du^{*}
}.
\]
The support of the measure $du^{*}$ is contained in 
the set $\{u=(0,u_{b})\}$. We use a notation
$\sigma=(\sigma_{a},\sigma_{b})\in{\RR}^{m}\times{\RR}^{d-m}$. 
If $Q(q,p,\varphi)=1$ then there exists a resolution map $w=g(u)$ such that 
$\sigma_{a}^{k}$ takes values both $+1$ and $-1$ 
in arbitrary local coordinate, hence 
\[
\sum_{\sigma_{a}\in S(m)}\sigma_{a}^{k}=0.
\]
It follows that 
\begin{eqnarray*}
\sum_{\sigma\in S(d)}
\sigma^{k}\xi_{n}(0,u_{b})
= \sum_{\sigma_{b}\in S(d-m)}\sigma_{b}^{k}\xi_{n}(0,u_{b})
\sum_{\sigma_{a}\in S(m)}\sigma_{a}^{k}=0,
\end{eqnarray*}
therefore, $\Theta=0$, which completes  Corollary \ref{Corollary:111}.
(Q.E.D.) 

\subsection{Proof Corollary \ref{Corollary:222}}

By using the optimal inverse temperature $\beta^{*}$, we define $T=1/(\beta^{*}\log n)$.
By the definition, ${\cal F}=\EE_{w}^{\beta^{*}}[nL_{n}(w)]$. By using 
Theorem \ref{Theorem:NC2001} and Theorem \ref{Theorem:main},
\[
\lambda\log n =T\lambda\log n
+U_{n}\sqrt{T\lambda (\log n) /2} +O_{p}(\log\log n),
\]
which is equivalent to
\[
T+\frac{U_{n}\sqrt{T}}{\sqrt{2\lambda\log n }} -1 +O_{p}\Bigl(\frac{\log\log n}{\log n}\Bigr)=0.
\]
Therefore,
\[
\sqrt{T}=1-\frac{U_{n}}{\sqrt{8\lambda \log n}}+o_{p}(\frac{1}{\sqrt{\lambda \log n}}),
\]
resulting that 
\[
\beta^{*}\log n =1+ \frac{U_{n}}{\sqrt{2\lambda \log n}}+o_{p}(\frac{1}{\sqrt{\lambda \log n}}),
\]
which completes Corollary \ref{Corollary:222}. (Q.E.D.)

\subsection{Proof of Corollary \ref{Corollary:333}}

By using Theorem \ref{Theorem:main},
\begin{eqnarray*}
\EE_{w}^{\beta_{1}}[nL_{n}(w)]&=&n L_{n}(w_{0})+\frac{\lambda}{\beta_{1}}
+O_{p}(\sqrt{\log n }), \\
\EE_{w}^{\beta_{2}}[nL_{n}(w)]&=&n L_{n}(w_{0})+\frac{\lambda}{\beta_{2}}
+O_{p}(\sqrt{\log n}).
\end{eqnarray*}
Since $(1/\beta_{1}-1/\beta_{2})=O_{p}(\log n)$, 
\[
\lambda=\frac{\EE_{w}^{\beta_{1}}[nL_{n}(w)]-\EE_{w}^{\beta_{2}}[nL_{n}(w)]}{1/\beta_{1}-1/\beta_{2}}
+O_{p}(1/\sqrt{\log n}), 
\]
which shows Corollary \ref{Corollary:333}. (Q.E.D.)

\subsection{Proof Theorem \ref{Theorem:regular}}

By using eq.(\ref{eq:LnKn}), eq.(\ref{eq:KnKn}), the proof of Theorem \ref{Theorem:regular} 
results in evaluating $E_{n}^{\beta}[nK_{n}(w)]$.  
By Lemma \ref{Lemma:start} for the case $r=1/4$, 
\begin{equation}\label{eq:ZZZ2}
E_{n}^{\beta}[nK_{n}(w)]= \frac{D_{n}+o_{p}(\exp(-\sqrt{n}))}{C_{n}+o_{p}(\exp(-\sqrt{n}))},
\end{equation}
where $C_{n}$ and $D_{n}$ are respectively defined by 
\begin{eqnarray}
C_{n}&=& \int_{K<1/n^{1/4}}\exp(-n\beta K_{n}(w))\varphi(w)dw, \label{eq:An3} \\
D_{n}&=& \int_{K<1/n^{1/4}}nK_{n}(w)\exp(-n\beta K_{n}(w))\varphi(w)dw. \label{eq:Bn3}
\end{eqnarray}
If a statistical model is regular, the maximum likelihood estimator $\hat{w}$ 
converges to $w_{0}$ in probability. 
Let $J_{n}(w)$ be $d\times d$ matrices defined by 
\[
(J_{n})_{ij}(w)= \frac{\partial^{2}K_{n}}{\partial w_{i}\partial w_{j}}(w).
\]
There exists a parameter $w^{*}$ such that 
\begin{eqnarray*}
K_{n}(w)&=&K_{n}(\hat{w})+\frac{1}{2}(w-\hat{w})\cdot J_{n}(w^{*})(w-\hat{w}).
\end{eqnarray*} 
Since $\hat{w}\rightarrow w_{0}$ in probability, $w^{*}\rightarrow w_{0}$ in
probability. Then
\begin{eqnarray*}
&& \|J_{n}(w^{*})-J(w_{0})\| \leq  \|J_{n}(w^{*})-J_{n}(w_{0})\|
+\|J_{n}(w_{0})-J(w_{0})\| \\
&& \leq \|w^{*}-w_{0}\|\sup_{K(w)<1/n^{1/4}}\Bigr{\|}
\frac{\partial J_{n}(w)}{\partial w}\Bigr{\|}
+\|J_{n}(w_{0})-J(w_{0})\|,
\end{eqnarray*}
which converges to zero in probability
as $n\rightarrow\infty$. Therefore
\[
J_{n}(w^{*})=J(w_{0})+o_{p}(1).
\]
Since a statistical model is regular, $J(w_{0})$ is a positive definite matrix. 
\begin{eqnarray*}
C_{n}&=&\exp(-n\beta K_{n}(\hat{w}))\\
&& \times \int_{K(w)<n^{1/4}}
\exp(-\frac{n\beta}{2}(w-\hat{w})\cdot (J(w_{0})+o_{p}(1))(w-\hat{w}))\varphi(w)dw.
\end{eqnarray*}
By substituting 
\[
u=\sqrt{n\beta}(w-\hat{w}),
\]
it follows that 
\begin{eqnarray*}
C_{n}&=&\exp(-n\beta K_{n}(\hat{w}))(n\beta)^{-d/2} \\
&& \times \int
\exp(-\frac{1}{2}u \cdot (J(w_{0})+o_{p}(1))u  )\varphi(\hat{w}+\frac{u}{\sqrt{n\beta}})du \\
&=&\frac{(2\pi)^{d/2}\exp(-n\beta K_{n}(\hat{w}))(\varphi(\hat{w})+o_{p}(1))}
{(n\beta)^{d/2}\det(J(w_{0})+o_{p}(1))^{1/2}}.
\end{eqnarray*}
By the same way, 
\begin{eqnarray*}
D_{n}&=&\exp(-n\beta K_{n}(\hat{w}))\\
&& \times \int_{K(w)<1/n^{1/4}}\Bigl(
nK_{n}(\hat{w})+\frac{n}{2}(w-\hat{w})\cdot (J(w_{0})+o_{p}(1))(w-\hat{w})\Bigr)\\
&&\times 
\exp\Bigl(-\frac{n\beta}{2}(w-\hat{w})\cdot (J(w_{0})+o_{p}(1))(w-\hat{w})\Bigr)\varphi(w)dw \\
&=&\frac{(2\pi)^{d/2}\exp(-n\beta K_{n}(\hat{w}))(\varphi(\hat{w})+o_{p}(1))}
{(n\beta)^{d/2}\det(J(w_{0})+o_{p}(1))^{1/2}}
\Bigl(nK_{n}(\hat{w})+\frac{d}{2\beta}+o_{p}(1)\Bigr).
\end{eqnarray*}
Here $nK_{n}(\hat{w})=O_{p}(1)$, because a true distribution is regular for a statistical model. 
Therefore, 
\[
\EE_{w}^{\beta}[nL_{n}(w)]=nL_{n}(w_{0})+nK_{n}(\hat{w})+\frac{d}{2\beta}+o_{p}(1),
\]
which completes Theorem \ref{Theorem:regular}. (Q.E.D.) 

\section{A Method How to Use WBIC}

In this section we show a method how to use WBIC in statistical model
evaluation. The main theorems have 
already been mathematically proved, hence $\mathrm{WBIC}$ has a
theoretical support. The following exeperiment was
conducted not for proving theorems but for illustrating a method how to use 
it. 

\subsection{Statistical Model Selection}

\begin{table}[tb]
\begin{center}
\begin{tabular}{|c|c|c|c|c|c|c|}
\hline
$H$ & 1 & 2 & 3 & 4 & 5 & 6 \\
\hline
$\mathrm{WBIC}_{1}$  Ave.  & 17899.8 & 3088.9 & 71.1 & 77.9 & 83.3 & 87.7 \\
\hline
$\mathrm{WBIC}_{1}$  Std.  & 1081.3 &227.0 & 3.7 &4.0  & 4.0 &4.2 \\
\hline
$\mathrm{WBIC}_{2}$  Ave. & 17828.7 & 3017.9& 0 & 6.8& 12.2 & 16.6 \\
\hline
$\mathrm{WBIC}_{2}$  Std.  & 1081.2 & 226.7 &0 & 1.8 & 2.3& 2.3 \\
\hline
\end{tabular}
\end{center}
\caption{WBIC in Model Selection}
\label{table:222}
\end{table}

Firstly, we study model selection by using WBIC. 

Let $x\in{\RR}^{M}$, $y\in {\RR}^{N}$, and $w=(A,B)$, where
$A$ is an $H\times M$ matrix and $B$ is an $M\times H$ matrix. 
A reduced rank regression model is defined by
\[
p(x,y|w)=\frac{r(x)}{(2\pi\sigma^{2})^{N/2}}\exp\Bigl(-\frac{1}{2\sigma^{2}}\|y-BAx\|^{2}\Bigr),
\]
where $r(x)$ is a probability density function of $x$ and $\sigma^{2}$ is the variance 
of an output. Let ${\cal N}_{M}(0,\Sigma)$ denote 
the $M$ dimensional normal distribution 
with the average zero and the covariance matrix $\Sigma$.

In an experiment, we set $\sigma=0.1$, 
$r(x)={\cal N}_{M}(0,3^{2}I)$, where $I$ is the identity matrix, 
and 
$
\varphi(x)={\cal N}_{d}(0,10^{2}I). 
$
The true distribution was fixed as $p(x,y|w_{0})$, where $w_{0}=(A_{0},B_{0})$ was
determined so that $A_{0}$ and $B_{0}$ were respectively 
an $H_{0}\times M$ matrix and an $M\times H_{0}$ matrix. 
Note that, in reduced rank regression models, RLCTs and multiplicities were 
clarified by \citep{Aoyagi} and $Q(K(w),\varphi(w))=1$ for
arbitrary $q(x)$, $p(x|w)$, and $\varphi(w)$. 
In the experiment, $M=N=6$ and 
the true rank was set as $H_{0}=3$. Each element of $A_{0}$ and $B_{0}$ was 
taken from ${\cal N}_{1}(0,0.2^2)$ and fixed. 
From the true distribution $p(x,y|w_{0})$, 100 sets of $n=500$ training samples were 
generated. 

The Metroplois method was employed for 
sampling from the posterior distribution, 
\[
p(w|X_{1},X_{2},...,X_{n})\propto \exp(-\beta nL_{n}(w)+\log \varphi(w)),
\]
where $\beta=1/\log n$. 
Every Metropolis trial was generated from 
a normal distribution ${\cal N}_{d}(0,(0.0012)^2I)$, by which 
the exchange probability was 0.3-0.5.
First 50000 Metropolis trails were not used. After 50000 trails, $R=2000$ parameters 
$\{w_{r};r=1,2,...,R\}$ were obtained in every 100 Metropolis steps. The expectation
value of a function $G(w)$ over the posterior distribution was approximated by
\[
\EE_{w}^{\beta}[G(w)]=\frac{1}{R}\sum_{r=1}^{R}G(w_{r}).
\]
The six statistical models $H=1,2,3,4,5,6$ were compared by the criterion, 
\[
\mathrm{WBIC}=\EE_{w}^{\beta}[nL_{n}(w)],\;\;\;(\beta=1/\log n).
\]
To compare these values among several models, we show both 
$\mathrm{WBIC}_{1}$ and $\mathrm{WBIC}_{2}$ in Table \ref{table:222}.
In the table, 
the average and the standard deviation of $\mathrm{WBIC}_{1}$ defined by 
\[
\mathrm{WBIC}_{1}=\mathrm{WBIC}- nS_{n},
\]
for 100 independent sets of training samples are shown, 
where the empirical entropy of the true distribution
\[
S_{n}=-\frac{1}{n}\sum_{i=1}^{n}\log q(X_{i})
\]
does not depend on a statistical model. Also 
$\mathrm{WBIC}_{2}$ in Table \ref{table:222} shows the 
average and the standard deviation of 
\[
\mathrm{WBIC}_{2}=\mathrm{WBIC}- \mathrm{WBIC}(3),
\]
where $\mathrm{WBIC}(3)$ is the WBIC for $H=H_{0}=3$.
In 100 independent sets of training samples, 
the true model $H=3$ was chosen 100 times in this experiment, 
which demonstrates a typical application method of WBIC. 

\subsection{Estimating RLCT}

Secondly, we study a method how to estimate an RLCT. 
By using the same experiment as the foregoing subsection, 
we estimated RLCTs of reduced rank regression models 
by using Corollary \ref{Corollary:333}.
Based on eq.(\ref{eq:lambda-estimated}), the
estimated RLCT is given by 
\[
\hat{ \lambda}=
\frac{
\EE_{w}^{\beta_{1}}[nL_{n}(w)]-\EE_{w}^{\beta_{2}}[nL_{n}(w)]
}{
1/\beta_{1}-1/\beta_{2}
},
\]
where $\beta_{1}=1/\log n$ and $\beta_{2}=1.5/\log n$ were
used and 
\[
\EE_{w}^{\beta_{2}}[nL_{n}(w)]
=\frac{
\EE_{w}^{\beta_{1}}[nL_{n}(w)\exp(-(\beta_{2}-\beta_{1})nL_{n}(w))]
}{
\EE_{w}^{\beta_{1}}[\exp(-(\beta_{2}-\beta_{1})nL_{n}(w))]
}.
\]
Theory $\lambda$ in Table \ref{table:333} shows the theoretical values of 
RLCTs of reduced rank regression. For 
the cases when true distributions are unrealizable by statistical models,
RLCTs are given by half the dimension of the parameter space, 
$
\lambda=H(M+N-H)/2.
$
In Table \ref{table:333}, averages and standard deviations of $\lambda$ 
shows estimated RLCTs. The theoretical RLCTs were well estimated. 
The difference between theory and experimental results was caused by
the effect of the smaller order terms than $\log n$. 
In the case the multiplicity $m=2$, the term $\log\log n$ also 
affected the results. 

\begin{table}[tb]
\begin{center}
\begin{tabular}{|c|c|c|c|c|c|c|}
\hline
$H$ & 1 & 2 & 3 & 4 & 5 & 6 \\
\hline
Theory $\lambda$ & 5.5 & 10 & 13.5 & 15 & 16 & 17 \\
\hline
Theory $m$ & 1 & 1 & 1 & 2 & 1 & 2 \\
\hline
Average $\lambda$ & 5.50 & 9.93  & 13.44 & 14.69 & 15.74 & 16.53 \\
\hline
Std. Dev. $\lambda$ & 0.19 & 0.32 & 0.47 & 0.60 & 0.66 & 0.88 \\
\hline
\end{tabular}
\end{center}
\caption{RLCTs for the case $H_{0}=3$}
\label{table:333}
\end{table}

\section{Discussion}

In this section, we discuss the widely applicable information criterion
 from three different points of view.

\subsection{WAIC and WBIC}

Firstly, let us study the difference between the free energy and the generalization error. 
In the present paper, we study the Bayes free energy
${\cal F}$ as the statistical model selection criterion. 
Its expectation value is given by 
\[
\EE[{\cal F}]=nS+
\int q(x^{n})\log\frac{q(x^{n})}{p(x^{n})}dx^{n},
\]
where $S$ is the entropy of the true distribution, 
\begin{eqnarray*}
q(x^{n})&=&\prod_{i=1}^{n}q(x_{i}),\\
p(x^{n})&=&\int \prod_{i=1}^{n}p(x_{i}|w)\varphi(w)dw,
\end{eqnarray*}
and $dx^{n}=dx_{1}dx_{2}\cdots dx_{n}$. 
Hence minimization of $\EE[{\cal F}]$ is equivalent to 
minimization of the Kullback-Leibler distance from the 
$q(x^{n})$ to $p(x^{n})$. 

There is a different model evaluation criterion, 
which is the generalization loss defined by
\begin{equation}\label{eq:GGG}
{\cal G}=-\int q(x)\log p^{*}(x) dx,
\end{equation}
where $p^{*}(x)$ is the Bayes predictive distribution defined by 
$
p^{*}(x)=\EE_{w}^{\beta}[p(x|w)],
$
with $\beta=1$. 
The expectation value of ${\cal G}$ satisfies
\[
\EE[{\cal G}]=S+\EE\Bigl{[}\int q(x)\log \frac{q(x)}{p^{*}(x)}dx\Bigr{]}.
\]
Hence minimization of $\EE[{\cal G}]$ is equivalent to 
minimization of the Kullback-Leibler distance from 
$q(x)$ to $p^{*}(x)$. 
Both of ${\cal F}$ and ${\cal G}$ are important in statistics and
learning theory, however, they are different criteria. 

The well-known model selection criteria AIC and BIC 
are respectively defined by 
\begin{eqnarray}
\mathrm{AIC}&=& L_{n}(\hat{w})+\frac{d}{n},\label{eq:WAICDEF} \\
\mathrm{BIC}&=& nL_{n}(\hat{w})+\frac{d}{2}\log n.
\end{eqnarray}
If a true distribution is realizable by and regular for
a statistical model, then 
\begin{eqnarray*}
\EE[\mathrm{AIC}]&=& \EE[{\cal G}]+o(\frac{1}{n}),\\
\EE[\mathrm{BIC}]&=& \EE[{\cal F}]+O(1).
\end{eqnarray*}
These relations can be generalized onto singular
statistical models. We define WAIC and WBIC by 
\begin{eqnarray*}
\mathrm{WAIC}&=& T_{n}+V_{n}/n, \\
\mathrm{WBIC}&=& \EE_{w}^{\beta}[nL_{n}(w)], \;\;\; \beta = 1/\log n,
\end{eqnarray*}
where 
\begin{eqnarray*}
T_{n}&=&-\frac{1}{n}\sum_{i=1}^{n}\log p^{*}(X_{i}|w), \\
V_{n}&=&\sum_{i=1}^{n}
\Bigl{\{}\EE_{w}[(\log p(X_{i}|w))^{2}]
- \EE_{w}[\log p(X_{i}|w)]^{2}\Bigr{\}}.
\end{eqnarray*}
Then, even if a statistical model is unrealizable by and 
singular for a statistical model, 
\begin{eqnarray}
\EE[\mathrm{WAIC}]&=& \EE[{\cal G}]+O(\frac{1}{n^{2}}),\label{eq:waic11}\\
\EE[\mathrm{WBIC}]&=& \EE[{\cal F}]+O(\log\log n),\label{eq:wbic11}
\end{eqnarray}
where eq.(\ref{eq:waic11}) was proved in \citep{Cambridge2009,JMLR2010}, whereas 
eq.(\ref{eq:wbic11}) has been proved in the present paper. 
Moreover, if a statistical model is realizable by and 
regular for a statistical model, WAIC and WBIC respectively
coincide with AIC and BIC, 
\begin{eqnarray*}
\mathrm{WAIC}&=& \mathrm{AIC}+o_{p}(\frac{1}{n}), \\
\mathrm{WBIC}&=& \mathrm{BIC}+o_{p}(1). 
\end{eqnarray*}
Theoretical comparison of WAIC and WBIC in 
singular model selection is the important problem for future study. 
\vskip5mm\noindent
{\bf Remark.} If a prior distribution is positive at the
optimal set of parameters, then RLCTs are smaller than $d/2$ in singular models,
resulting that
both WAIC and WBIC in singular models are respectively smaller than AIC and BIC.
Theorefore, if Bayes estimation is applied to singular models, a larger model can be employed 
with a smaller generalization error. If a true model is unrealizable by 
any finite size model, this is a good property from the viewpoint of 
the best balance of bias and variance, however, this fact simultaneously
means the weaker consistency in model selection. If a true model is realizable by
some finite size model, and if the main purpose of statistical model evaluation is to find 
the true model, Jeffreys' prior is recommended. Note that Jeffreys' prior is equal to zero
at singularities and $\lambda\geq d/2$ holds \citep{Cambridge2009}. 
However, Jeffreys' prior is not appropriate to a case when a true distribution is
unrealizable by a finite model. It is well known in statistics and learning theory 
that consistency in model selection is different
from minimization of the generalization error.

\subsection{Other Methods How to Evaluate Free Energy}

Secondly, we discuss several methods how to numerically 
evaluate the Bayes free energy. 
There are three methods other than WBIC. 

Firstly, 
let $\{\beta_{j};j=0,1,2,...,J\}$ be a sequence
which satisfies
\[
0=\beta_{0}<\beta_{1}<\cdots<\beta_{J}=1.
\]
Then the Bayes free energy satisfies 
\[
{\cal F}=-\sum_{j=1}^{J}\log \EE_{w}^{\beta_{j-1}}
[\exp(-n(\beta_{j}-\beta_{j-1})L_{n}(w))].
\] 
This method can be used without asymptotic theory. We can 
estimate ${\cal F}$, if the number $J$ is sufficiently large and if 
all expectation values over 
the posterior distributions $\{\EE_{w}^{\beta_{j-1}}[\;\;]\}$ 
are precisely calculated. The disadvantage of this
method is its huge computational costs for accurate 
calculation. In the present paper, this method is referred to as 
`all temperatures method'.  

Secondly, the importance sampling method is 
often used. Let $H(w)$ be a function which approximates $nL_{n}(w)$. 
Then, for an arbitrary function $G(w)$, we define an expectation value
$\hat{\EE}_{w}[\;\;]$ by 
\[
\hat{\EE}_{w}[G(w)]=\frac{\int G(w)\exp(-H(w))\varphi(w)dw}
{\int \exp(-H(w))\varphi(w)dw}.
\]
Then 
\begin{eqnarray*}
{\cal F}&=&-\log \hat{\EE}_{w}[\exp(-nL_{n}(w)+H(w))] \\
&& -\log \int \exp(-H(w))\varphi(w)dw, 
\end{eqnarray*}
where the last term is the free energy of $H(w)$. 
Hence if we find $H(w)$ whose free energy is analytically
calculated and if it is easy to generate random samples from
$\hat{\EE}_{w}[\;\;]$, then ${\cal F}$ can be numerically evaluated. 
The accuracy of this method strongly depends on the choice of $H(w)$. 

Thirdly, a two-step method was proposed by \citep{Drton2}. 
Assume that we have theoretical values about RLCTs for all cases
about true distribution and statistical models. Then,
in the first step, a null hypothesis model is chosen by using
BIC. In the second step, 
the optimal model is chosen by using RLCTs with the assumption
that the null hypothesis model is a true distribution.
If the selected model is different from the null hypothesis model, 
then the same procedure is recursively applied until the null hypothesis model
becomes the optimal model. In this method, asymptotic theory is necessary but
RLCTs do not contain fluctuations because they are theoretical values. 

Compare with these three methods, WBIC needs 
asymptotic theory but not theoretical values of RLCTs. 
Moreover, $\mathrm{WBIC}$ can be used even if a
true distribution is unrealizable by a statistical model. 
The theoretical comparison of these four methods is
shown in Table \ref{table:444}. 

The effectiveness of a model selection method strongly depends on 
a statistical condition which is determined by 
a true distribution, a statistical model, a prior distribution, and 
a set of training samples.
Under some condition, one method may be more effective, 
however, under the other condition, another may be. 
The proposed method WBIC gives a new approach in numerical 
calculation of the Bayes free energy which is more useful with 
cooperation with the conventional method. 
It is a future study to clarify which method is recommended 
in what statistical conditions.

\begin{table}[tb]
\begin{center}
\begin{tabular}{|c|c|c|c|c|}
\hline
Method & Asymptotics & RLCT & Comput. Cost \\
\hline
All Temperatures &  Not used & Not Used & Huge \\
\hline
Importance Sampling & Not used & Not Used & Small \\
\hline
Two-Step & Used & Used & Small\\
\hline
WBIC & Used & Not Used & Small \\
\hline
\end{tabular}
\end{center}
\caption{Comparison of Several Methods}
\label{table:444}
\end{table}

\subsection{Algebraic geometry and Statistics} 

Lastly, let us discuss a relation between algebraic geometry and statistics. 
In the present paper, we define the parity of a statistical $Q(K(w),\varphi(w))$ and proved that
it affects the asymptotic behavior of WBIC. In this subsection we show three mathematical properties
of the parity of a statistical model. 

Firstly, the parity has a relation to the analytic continuation of $K(w)^{1/2}$.
For example, by using blow-up,
$
(a,b)=(a_{1},a_{1}b_{1})=(a_{2}b_{2},b_{2}),
$
it follows that analytic continuation of $(a^{2}+b^{2})^{1/2}$ is given by 
\[
(a^{2}+b^{2})^{1/2}=a_{1}\sqrt{1+b_{1}^{2}}=b_{2}\sqrt{a_{2}^{2}+1},
\]
which takes both positive and negative values. On the other hand,
$(a^{4}+b^{4})^{1/2}$ takes only nonnegative value. The parity 
indicates such difference.

Secondly, the parity has a relation to statistical model with a
restricted parameter set. For example, a statistical model
\[
p(x|a)=\frac{1}{\sqrt{2\pi}}\exp(-\frac{(x-a)^{2}}{2})
\]
whose parameter set is given by $\{a\geq 0\}$ is equivalent to a statistical model 
$p(x|b^{2})$ and $\{b\in {\RR}\}$. In other words, a statistical model which
has restricted parameter set is statistically equivalent to another even model which
has unrestricted parameter set. We have a conjecture that an even statistical model
has some relation to a model with a restricted parameter model. 

And lastly, the parity has a relation to the difference of 
$K(w)$ and $K_{n}(w)$. 
As is proven in \citep{NC2001}, the relation
\[
-\log\int\exp(-nK_{n}(w))\varphi(w)dw =
-\log\int\exp(-nK(w))\varphi(w)dw +O_{p}(1)
\]
holds independent of the parity of a statistical model. 
On the other hand, if $\beta=1/\log n$, then 
\begin{eqnarray*}
\EE_{w}^{\beta}[nK_{n}(w)]&=&
\frac{\int nK(w)\exp(-n\beta K(w))\varphi(w)dw}
{\int\exp(-n\beta K(w))\varphi(w)} \\
& & + U_{n}\sqrt{\log n} + O_{p}(1).
\end{eqnarray*}
If the parity is odd, then 
$U_{n}=0$, otherwise $U_{n}$ 
is not equal to zero in general. This fact shows that 
the parity shows difference in a fluctuation of the likelihood function.

\section{Conclusion}

We proposed a widely applicable Bayesian information
criterion (WBIC) which can be used 
even if a true distribution is unrealizable by
and singular for a statistical model and proved that WBIC has the same asymptotic 
expansion as the Bayes free energy. 
Also we developed a method how to estimate
real log canonical thresholds even if a true distribution is unknown. 

\subsection*{Acknowledgement}
This research was partially supported by the Ministry of Education,
Science, Sports and Culture in Japan, Grant-in-Aid for Scientific
Research 23500172.

\end{document}